\pdfoutput=1

\documentclass[11pt]{article}

\usepackage[final]{acl}

\usepackage{times}
\usepackage{latexsym}
\usepackage{colortbl}
\usepackage{listings}
\usepackage{makecell} 
\usepackage[T1]{fontenc}

\usepackage[utf8]{inputenc}

\usepackage{microtype}
\usepackage{xspace}
\usepackage{inconsolata}

\newcommand{\ie}{\emph{i.e.,}\xspace}

\newcommand{\eg}{\emph{e.g.,}\xspace}

\usepackage{graphicx}

\usepackage[utf8]{inputenc} 
\usepackage[T1]{fontenc}    
\usepackage{hyperref}       
\usepackage{url}            
\usepackage{booktabs}       
\usepackage{amsfonts}       
\usepackage{nicefrac}       
\usepackage{microtype}      
\usepackage{xcolor}         
\usepackage{amsmath}
\usepackage{amssymb}
\usepackage{nomencl}
\usepackage{glossaries}
\usepackage{xspace}
\usepackage{multirow}
\usepackage{bbding}
\usepackage{pifont}
\usepackage{algorithm}
\usepackage{subfigure}
\usepackage{enumitem}
\usepackage{multicol}
\usepackage{caption}
\usepackage{array}
\usepackage{fp}
\usepackage{makecell}
\usepackage{flushend}
\usepackage{cases}
\usepackage{color}
\usepackage{algpseudocode}
\usepackage{listings}
\usepackage{mathrsfs}
\usepackage{longtable}
\usepackage{colortbl}
\usepackage{bm}

\definecolor{codegreen}{rgb}{0,0.3,0.6}
\definecolor{codegray}{rgb}{0.5,0.5,0.5}

\newcommand{\ignore}[1]{}

\usepackage{tcolorbox}
\definecolor{darkorange}{RGB}{255, 140, 0}
\definecolor{lightgreen}{RGB}{145, 204, 117}
\definecolor{lightyellow}{RGB}{250, 200, 88}
\definecolor{lightred}{RGB}{238, 102, 102}
\definecolor{lightblue}{RGB}{115, 192, 222}
\newtcolorbox{promptbox}[2][Prompt]{
colback=black!5!white,
arc=5pt, 
boxrule=0.5pt,
fonttitle=\bfseries,
title=#1, 
before upper={\scriptsize}, fontupper=\fontfamily{ptm}\selectfont,
colframe=#2, 
}

\title{RV-Syn: Rational and Verifiable Mathematical Reasoning \\Data Synthesis Based on Structured Function Library}

\author{
    \textbf{
        Jiapeng Wang\textsuperscript{\rm{1,3}\thanks{\ \ Equal contribution.}},
        Jinhao Jiang\textsuperscript{\rm{1,3}\footnotemark[1]},
    } \\
    \textbf{
        Zhiqiang Zhang\textsuperscript{\rm{2}\ },
        Jun Zhou\textsuperscript{\rm{2}\ },
        Wayne Xin Zhao\textsuperscript{\rm{1,3}\thanks{\ \ Corresponding author.}},
    } \\
    \textsuperscript{1}Gaoling School of Artificial Intelligence, Renmin University of China  
    \textsuperscript{2}Ant Group \\
    \textsuperscript{3}Beijing Key Laboratory of Research on Large Models and Intelligent Governance \\
    \texttt{wangjp1010@ruc.edu.cn, batmanfly@gmail.com}
}

\begin{document}
\maketitle
\begin{abstract}
The advancement of reasoning capabilities in Large Language Models (LLMs) requires substantial amounts of high-quality reasoning data, particularly in mathematics. Existing data synthesis methods, such as data augmentation from annotated training sets or direct question generation based on relevant knowledge points and documents, have expanded datasets but face challenges in mastering the internal logic of the problem during generation and ensuring the verifiability of the solutions. To address these issues, we propose \textbf{RV-Syn}, a novel \underline{\textbf{R}}ational and \underline{\textbf{V}}erifiable mathematical \underline{\textbf{Syn}}thesis approach. 
RV-Syn first constructs a structured library of mathematical operations and then composes them into executable computational graphs, which serve as verifiable solution blueprints. These graphs are subsequently back-translated into complex problems, enabling solution-guided, logic-aware problem generation while inherently ensuring the verifiability of the solving process. 
Experimental results show RV-Syn surpasses existing synthesis methods, including those involving human-crafted problems.
Our method achieves a 6.3\% performance gain over the previous state-of-the-art synthetic data on LLaMA-3-8B and demonstrates superior data efficiency, outperforming others with only half the training data (50k vs. 100k), enabling a more scalable and robust reasoning dataset generation framework.
\end{abstract}

\section{Introduction}
The development of advanced reasoning Large Language Models (LLMs)~\cite{survey,openai2024reasoning} has markedly improved their ability to address complex tasks across domains such as mathematics, science, and coding. This highlights the importance of synthesizing complex reasoning data to drive further advancements, given the limited availability of high-quality annotated instructions~\cite{AI-Assisted,scaling}.

To address this scarcity, researchers have explored various synthesizing methods, particularly in the mathematics domain. The mainstream methods involve data augmentation based on existing annotated training sets, such as GSM8K~\cite{gsm8k} and MATH~\cite{dan2021math}, ranging from self-evolving instructions~\cite{wizardlm,Automatic_Instruction_Evolving} and question paraphrasing~\cite{metamath}, to solution augmentation~\cite{mathgenie}. However, these methods are limited by the available training data, constraining the synthesis diversity~\cite{scratch,DBLP:journals/corr/abs-2310-05506}. To enhance diversity, recent approaches enable LLMs to generate a large scale of questions from various mathematics-related sources, including web pages~\cite{mammoth2} and knowledge points~\cite{mathscale} from web corpora or textbooks. 
\begin{figure*}[ht]
    \centering    \includegraphics[width=0.95\textwidth]{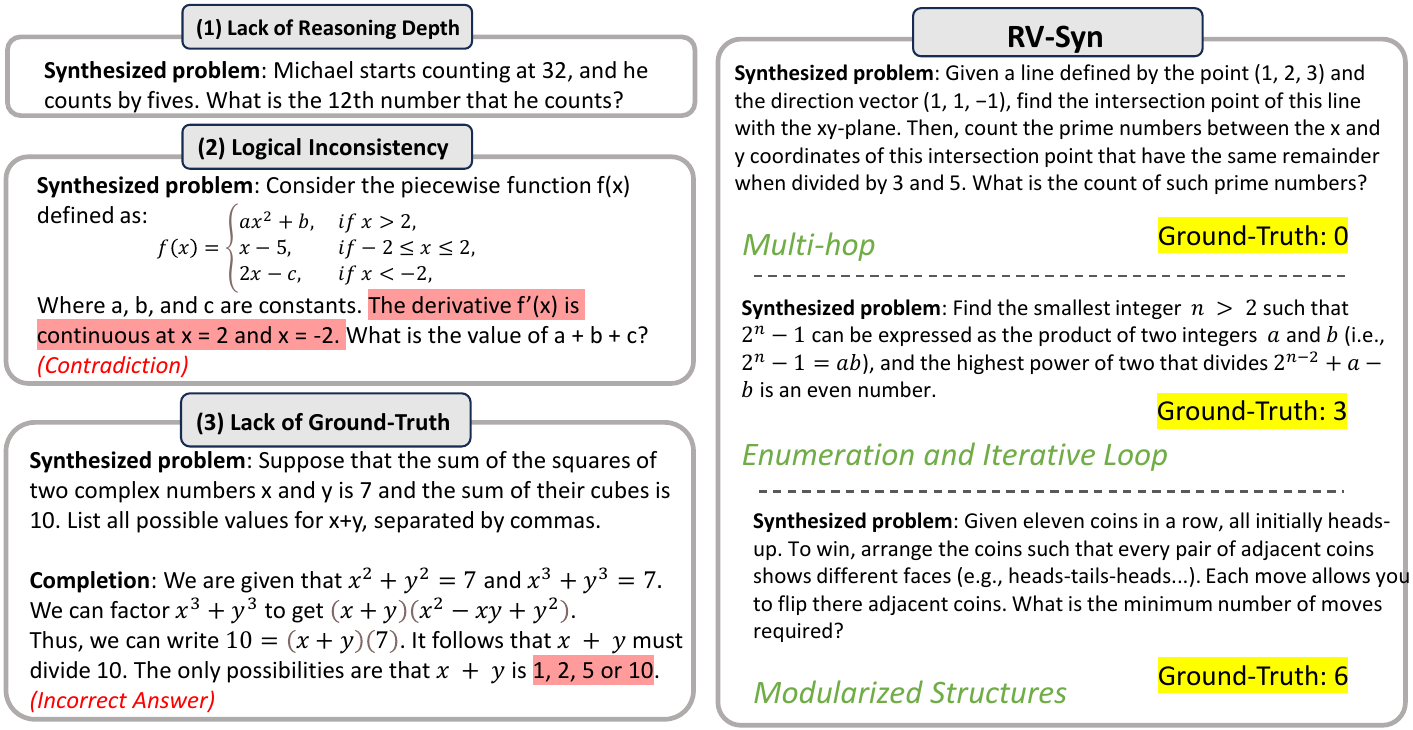}
  \caption{\textbf{(Left)} Illustration of the limitations of existing synthesis methods; \textbf{(Right)} Examples of data synthesized by RV-Syn, which naturally support sophisticated reasoning structures and provide ground-truth answers.}
  \label{fig:example}
\end{figure*}

However, as shown in Figure \ref{fig:example}, these direct problem generation methods suffer from a fundamental limitation. We argue this stems from the auto-regressive nature of LLMs, which leads to a lack of forward-looking planning during generation. When composing a math problem, a model might introduce entities and quantities without a coherent plan, frequently resulting in internal contradictions, unsolvable scenarios, or trivial questions, particularly in complex, multi-step reasoning problems. Furthermore, this approach makes it difficult to validate the correctness of the generated solutions, often compromising the quality and reliability of the training data.

To address the aforementioned challenges, we draw inspiration from how human educators create problems. Just as one cannot \textit{directly write down an Olympic-level math problem without deep consideration and careful curation of the underlying problem-solving process}, the same holds true for models.
Specifically, human educators obtain abstract independent computation goals from historical experience, such as ``projecting a vector onto a plane'', and ``solving for the smallest prime number that satisfies specific conditions''. Then, when designing a new question, they would combine these computation goals to obtain a new solution or introduce them into an existing solution. Finally, they derive the final problem based on this elaborated solution process.
This strategy ensures the development of problems with coherent and logically consistent solution processes, thereby minimizing the internal inconsistencies of the final problems.

Inspired by this, we propose \textbf{RV-Syn}, a novel framework for mathematical problem synthesis that emphasizes enhanced rationale and verifiability. Instead of direct generation, RV-Syn first synthesizes a structured computational graph. This graph serves as a formal blueprint, explicitly defining the problem’s reasoning structure and guaranteeing a valid, executable solution path. Subsequently, this logically consistent blueprint is back-translated into a natural language problem.
By generating questions through solution guidance, RV-Syn effectively provides the rich semantics and control flows required for complex reasoning. As shown in Figure~\ref{fig:example}, this enables the synthesis of diverse and sophisticated reasoning structures—including multi-hop reasoning, iterative loops, and modular structures—while naturally providing verifiable ground-truth labels.

We conduct extensive experiments to evaluate the proposed RV-Syn method using various LLMs. The results demonstrate that our method achieves superior performance across five benchmarks compared to existing approaches, {including those involving human-designed problems}~(\eg Numina-Math). Notably, RV-Syn outperforms the previous state-of-the-art method while utilizing only half the data, leading to a more efficient scaling curve. 





\section{Related Work}

\paragraph{Mathematical Reasoning.}
 Researchers have proposed various approaches to enhance the mathematical reasoning capabilities, including methods applied in training or inference stages. 
During the training stage, existing methods aim to enhance the LLMs from the aspects of pre-training and post-training. Specifically, some studies~\cite{llemma,Shao2024DeepSeekMath,syne} involve collecting extensive math-related corpora and enhancing the foundational mathematical capabilities of LLMs through continual pre-training. In contrast, other studies~\cite{DBLP:conf/emnlp/ChenWCS024,dart,MUSTARD,openmathinstruct} focus on synthesizing a substantial amount of high-quality math-related instructional data, further refining the problem-solving abilities of LLMs through post-training. 
During the inference stage, there are two primary prompting approaches, including Chain-of-Thought~(CoT)~\cite{wei2023chain} and Program-of-Thought~(PoT)~\cite{pot}. Specifically, some studies~\cite{react,reflexion,least_to_most} utilize CoT to elicit LLMs' inherent reasoning ability with a step-by-step reasoning process. Furthermore, other studies~\cite{tora, mathcoder,DBLP:journals/corr/abs-2401-05384, codeplan} incorporate PoT to allow LLMs to utilize external computers during reasoning.
In this work, we focus on the training stage and try to synthesize instruction data by leveraging the strengths of both the CoT and PoT methods. 

\paragraph{Data Synthesis for Math Instruction.}
Existing research has proposed various synthesizing approaches:
Firstly, existing studies focus on synthesizing new mathematical problems based on few-shot problems~\cite{xwin}, math-related documents~\cite{mammoth2}, key knowledge points~\cite{kpmath}, or self-evolution~\cite{wizardmath}. Furthermore, with the improvement of LLMs' capabilities, subsequent studies~\cite{jiuzhang, scalequest}  utilizingexplore the utilization of open-source LLMs~(\ie DeepSeek-Math and Qwen-Math) to scale the number of synthesized problems.
Despite their effectiveness, direct problem generation suffers from a lack of fine-grained control over the problem's internal logic, leading to lower-quality problems. Besides, these methods frequently lack mechanisms for validating the correctness of generated answers, which may introduce errors and affect the final performance.
In contrast, our method starts from the internal logic of problem-solving, rather than directly synthesizing problems. This provides more refined logical consistency and improves problem quality, while also enabling effective correctness control.

\begin{figure*}[ht]
    \centering    \includegraphics[width=1\textwidth]{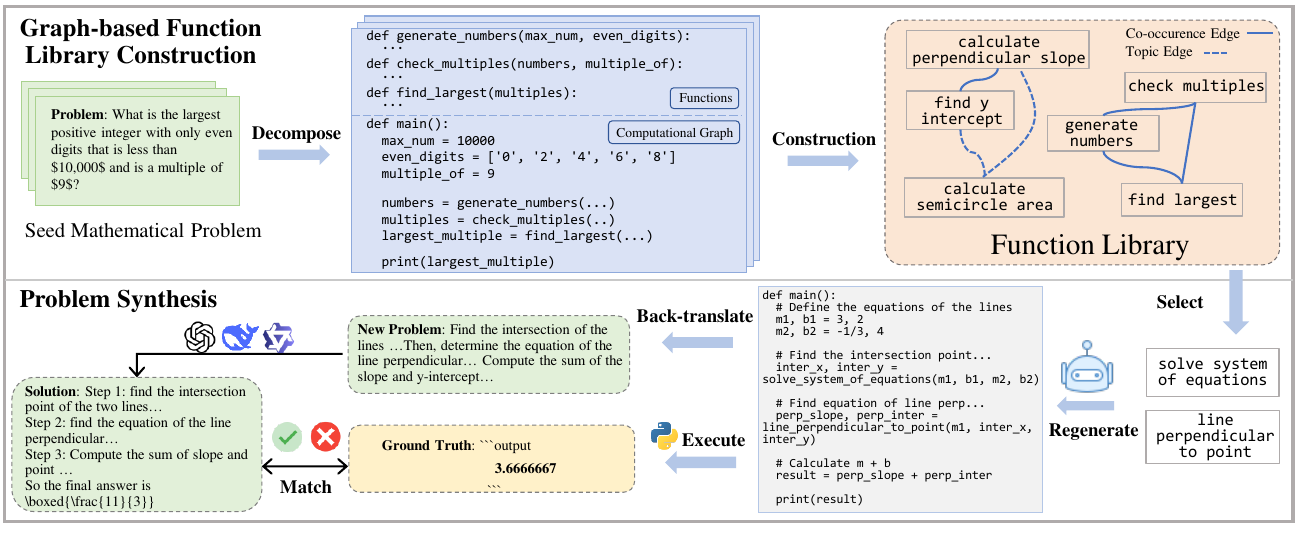}
    \vspace{-0.3cm}
  \caption{The pipeline of our proposed method. The upper part demonstrates the construction of the operation function library. The lower part illustrates the synthesis process.}
  \label{fig:pipeline}
\end{figure*}

\section{Approach}

In this section, we provide a comprehensive introduction to our proposed RV-Syn method. Its essence involves developing a comprehensive set of mathematical operations and skills. These skills are then systematically combined to construct the solution process, ultimately enabling the generation of complex problems through back-translation. The complete pipeline comprises three core stages: (1) decomposing existing seed problems to generate a set of computational graphs (Section \ref{decompose}), (2) extracting functions from these computational graphs and constructing a graph-format function library (Section \ref{func library}), and (3) combining selected functions to reconstruct new computational graphs, which are subsequently back-translated into problems (Section \ref{compose}). The overall data flow of our method is illustrated in Figure \ref{fig:pipeline}.

\subsection{Problem Decomposition}\label{decompose}
We first represent the solutions of existing mathematical problems with sequences of functions into a standardized format shown in Appendix~\ref{appendix:format} leveraging LLMs. Each function represents a specific mathematical skill with defined parameters, forming an executable \textit{computational graph} that captures the data flow of the problem-solving process (examples shown in Figure~\ref{fig:pipeline}).
To ensure the correctness of the extracted solutions and their associated functions, we filter out erroneous cases by executing the solution code using a Python interpreter and comparing its output with the ground-truth answer provided in the corresponding Chain-of-Thought annotation. This results in a large-scale, high-quality set of computational graphs derived from real-world problems.

\subsection{Graph-based Function Library Construction}\label{func library}
After decomposing the problem, we extract functions from the computational graphs.
To further enhance the quality of the function library, we employ model-based labeling. We utilize LLMs to validate the correctness of the functions and identify their corresponding mathematical topics, eliminating any functions that contain errors. The prompts are shown in Appendix~\ref{appendix:prompts}.
To describe the relationships between these functions, each function is represented as a node, and they are organized into a graph-based function library.
Specifically, the construction process involves two key steps: structural and semantic-aware function merging, followed by co-occurrence and topic-based node connecting. 

\subsubsection{Structural and Semantic-aware Function Merging}

As previously discussed, each extracted function primarily encapsulates a specific mathematical operation or skill. Consequently, different functions may exhibit similarities in semantics (\ie possessing similar skills) or structure (\ie sharing similar computational logic). To preserve the diversity of function expressions while minimizing redundancy within the final function library, we merge functions that are similar in either structure or semantic aspects into one node. Therefore, the final node is a set that contains one or more similar nodes. In the subsequent description, we will take this setting as the default.

Specifically, we examine the similarity between two functions by analyzing both structure and semantic aspects. {For the structure, we focus on the computational operations without considering specific variable names. For instance, expressions such as ``\((a + b) / c\)'' and ``\((m + n) / d\)'' are considered equivalent.} To accomplish this, we extract the function body and parse it using Abstract Syntax Trees (ASTs). For the semantic, we compute hash values for the function's docstrings. If the ASTs or hash values of two functions are identical, we merge these functions into a single node. We show an example of two similar functions in Table~\ref{tab:similar} at Appendix~\ref{sec:simi}. This approach allows similar functions (e.g., matrix multiplication and matrix projection) to preserve their distinct mathematical meanings while minimizing redundancy in the final function library, offering advantages over direct deduplication. Our structure-based method using ASTs proves highly effective in practice, successfully identifying and merging approximately 28\% of redundant functions from the initial library.


\subsubsection{Co-occurrence and Topic-based Node Connecting}

After obtaining the final set of nodes through function merging, we further establish connections between relevant nodes to construct the final graph. We define two types of edges: co-occurrence-based edges and topic-consistent edges.

Specifically, for any two given nodes (\ie two sets of functions), we first check whether there exists at least one pair of functions—one from each set—that co-occurred in the same computational graph. If so, we add a co-occurrence-based edge between these two nodes.
If not, we then check whether there is at least one pair of functions sharing the same topic. If this condition is met, we add a topic-consistent edge.
If neither condition is satisfied, no edge is added between the nodes.

In this way, we finally obtain an incompletely connected graph that contains multiple types of edges, which depicts the relationships between nodes. This serves as the basis for the subsequent recombination of functions.

\subsection{Problem Synthesis}\label{compose}
Based on the constructed graph-format function library, we are able to synthesize complex problems in a solution-guided approach. This process consists of two primary steps: first sampling various functions to regenerate the computational graph, then synthesizing the final complex problems.

\subsubsection{Computational Graph Regeneration}\label{ft}
To balance the reasonableness and novelty of the computational graphs, we design three sampling strategies: first, selecting nodes connected by co-occurrence-based edges; second, selecting nodes connected by topic-based edges; and third, selecting unconnected nodes. During the sampling process, nodes are randomly selected using these strategies. Since each node represents a set of functions, we further randomly select one function per node.
Subsequently, {we generate a computational graph based on these selected functions} as detailed in Appendix~\ref{appendix:prompts}.
This approach offers several advantages: Firstly, by combining functions, the generated computational graph can comprehensively cover mathematical skills while minimizing redundancy. Secondly, the sampling strategy effectively balances high-frequency combinations of mathematical skills (i.e., connected by edges) with long-tail combinations (i.e., without connected edges). In our setup, we randomly select from the three strategies when synthesizing each data sample.

\subsubsection{Problem Back-translation and Verification}
The final stage of our pipeline generates a complete (problem, solution) pair and ensures its correctness, highlighting the core verifiability of our framework. First, we execute all newly generated computational graphs using an interpreter. Graphs that fail to execute due to errors are discarded. This step serves two key purposes: it validates the feasibility of the synthesized reasoning process and yields a reliable, ground-truth final answer from successful execution. 
With a verified computational graph and its ground-truth answer, we then back-translate the graph into a coherent natural-language math problem and generate its Chain-of-Thought (CoT) solution. Unlike direct generation methods, which struggle to verify the correctness of their outputs, our framework can compare the final answer derived from the LLM-generated CoT solution against the ground-truth answer obtained from executing the computational graph. Any (problem, solution) pair with mismatched answers is automatically discarded.
This execution-based filtering acts as a precise and automatic quality control mechanism, effectively eliminating instances of flawed reasoning or incorrect calculations in the final LLM-generated solutions.
We provide more statistics of the synthesis pipeline in Appendix \ref{app:statistic}.

\begin{table}[t]
    \resizebox{0.48\textwidth}{!}{
            \begin{tabular}{l | c | c c c }
			\toprule
			\textbf{Method} & \textbf{Category} & \textbf{\makecell{Open-source \\ Problem Crafter}} & \textbf{\makecell{Rational Problem \\Synthesize}} & \textbf{\makecell{Automatic Solution \\Verification}}  \\
			\midrule
			MetaMath & Aug & \ding{55} & \ding{55} & \ding{55} \\
                Orca-Math & Aug & \ding{55} & \ding{55} & \ding{55} \\
                WizardMath & Aug & \ding{55} & \ding{55} & \ding{55} \\
                MathGenie & Aug & \checkmark & \checkmark & \ding{55} \\
			Mammoth2 & New & \checkmark & \ding{55} & \ding{55} \\
                Jiuzhang3.0 & New & \checkmark & \ding{55} & \ding{55} \\
                ControlMath & New & \ding{55} & \checkmark & \ding{55} \\
                MathScale & New & \ding{55} & \ding{55} & \ding{55} \\
                PromptCoT & New & \checkmark & \checkmark & \ding{55} \\
                ScaleQuest & New & \checkmark & \ding{55} & \ding{55} \\
                \midrule
                RV-Syn & New & \checkmark & \checkmark & \checkmark \\
    		\bottomrule
		\end{tabular}
        }
        \caption{Comparison of Different Methods. \emph{Category} specifies whether the method augments existing data (``Aug'') or synthesizes new questions (``New''). \emph{Open-source Problem Crafter} indicates whether the method utilizes open-source models (\checkmark) or proprietary models like GPT-4 (\ding{55}) for problem generation. \emph{Rational Problem Synthesize} denotes whether the synthesized problems incorporate internal problem-solving logic. {\emph{Automatic Solution Verification}} denotes the method's ability to automatically verify the correctness of synthesized data.}
        \label{tab:discuss}
\end{table}
\subsection{Comparison to Previous Work}
We give a comparison in Table~\ref{tab:discuss}. The first line of research, including MetaMath~\cite{metamath}, Orca-Math~\cite{orcamath}, WizardMath~\cite{wizardmath}, MuggleMath~\cite{mugglemath}, MathGenie~\cite{mathgenie}, primarily focuses on augmenting existing problems or solutions, which leads to new instructions that are too similar to the original ones, thus limiting diversity.
Another line of research, represented by Mammoth2~\cite{mammoth2}, Jiuzhang3.0~\cite{jiuzhang}, KPMath~\cite{kpmath}, and ScaleQuest~\cite{scalequest}, synthesizes brand new mathematical problems based on math-related documents, knowledge points, or from scratch, offering greater diversity and scalability.
However, many of these methods overlook the internal logical processes when directly synthesizing problems.
Methods like ControlMath~\cite{DBLP:conf/emnlp/ChenWCS024} and PromptCoT~\cite{promptcot} attempt to enhance the rationality of synthesis by leveraging equation generators or injecting intermediate thought processes.
Nevertheless, they still lack mechanisms to verify the correctness of the synthesized data. Our RV-Syn method approaches problem generation from the perspective of internal problem-solving logic rather than directly producing problem statements without methodical consideration, thereby enhancing problem quality. Additionally, through executable computation graphs, our method enables automatic verification of solution correctness.

\begin{table*}[t]

    \label{tab:main}
    \centering
    \resizebox{\textwidth}{!}{
    \begin{tabular}{lccccccc}
        \toprule
        \textbf{Method} & \textbf{Problem Crafter} & \textbf{MATH-500} & \textbf{GSM8K} & \textbf{GSM-Hard} & \textbf{College Math} & \textbf{Olympiad Bench} & \textbf{Avg (Rel. Imp.)} \\
        \midrule
        \multicolumn{8}{c}{\textit{Models based on LLaMA-3-8B-Instruct}} \\
           Official Model  &       -      & 28.4 & 75.1  & 35.6  & 21.2  & 7.3  & 33.50 \\
        MetaMath     & ChatGPT            & 42.4  & 83.1  & 36.2  & 22.3  & 10.1  & 38.82 (+15.9\%) \\
        Orca-Math     & GPT-4            & 29.4  & \textbf{86.6}  & 42.7  & 19.8  & 11.1  & 37.92 (+13.2\%)\\
        Mammoth2     & 72B Model           & 42.2  & 78.6  & 40.3  & \textbf{33.0}  & 13.6  & 41.54 (+24.0\%)\\
        Jiuzhang3.0  & 7B Math Model       & 43.4  & 81.3  & 40.5  & 29.6  & 15.9  & 42.14 (+25.8\%)\\
        MathScale    & ChatGPT        & 43.4  & 81.3  & 38.4  & 32.6  & 13.6  & 41.86 (+25.0\%)\\
        PromptCoT    & 72B Model        & 43.4  & 78.2  & 41.5  & 25.8  & 15.7  & 40.92 (+22.1\%)\\
        ScaleQuest   & 7B Math Model       & 45.8  & 83.6  & 41.9  & 29.4  & 13.3  & 42.80 (+27.8\%)\\
        \midrule
        \rowcolor{lightyellow} RV-Syn         &       7B / 72B model         & \textbf{50.4}  & 82.6  & \textbf{44.5}  & 30.7  & \textbf{16.4}  & \textbf{44.92 (+34.1\%)} \\
        \bottomrule
        \multicolumn{8}{c}{\textit{Models based on Qwen2.5-7B-Instruct}} \\
           Official Model  &       -      & 74.0 & 86.8  & 59.9  & 44.1  & 36.0  & 60.16 \\
        MetaMath     & ChatGPT            &  74.0 & 90.9  &  67.6 & 45.5  &  35.0 & 62.60 (+4.1\%) \\
        Orca-Math     & GPT-4            & 74.2  & \textbf{92.0}  &  67.4 & 45.3  & 35.4  & 62.86 (+4.5\%) \\
        Mammoth2     & 72B Model           & 75.0  & 90.2  & 68.2  & 45.9  & 34.7  & 62.80 (+4.4\%) \\
        Jiuzhang3.0  & 7B Math Model       & 74.8  & 91.3  & 67.4  & 45.8  & 34.7  & 62.80 (+4.4\%) \\
        MathScale    & ChatGPT        & 74.0  & 91.0  & 67.6  & 45.3  & \textbf{36.4}  & 62.98 (+4.5\%) \\
        PromptCoT    & 72B Model        & 75.8  & 90.5  & 67.7  & 45.8  & 34.8  & 62.92 (+4.6\%)\\
        ScaleQuest   & 7B Math Model       & 75.0  & 91.4  & 67.6  & 45.9  & 34.4  & 62.86 (+4.5\%) \\
        \midrule
        \rowcolor{lightyellow} RV-Syn         &       7B / 72B model         & \textbf{76.8}  & 91.3  & \textbf{69.6}  & \textbf{46.1}  & \textbf{36.4}  & \textbf{64.04 (+6.4\%)} \\
        \bottomrule
        \multicolumn{8}{c}{\textit{Models based on Phi-3-mini}} \\
           Official Model  &       -      & 43.0 & 73.3  & 54.1  & 35.7  & 15.7  & 44.40 \\
        MetaMath     & ChatGPT            & 55.8  &  89.8 &  64.5 &  39.6 & 20.7  & 54.08 (+21.8\%) \\
        Orca-Math     & GPT-4            &  55.2 & \textbf{90.4}  & \textbf{65.3}  & 40.0  & 19.6  & 54.10 (+21.9\%) \\
        Mammoth2     & 72B Model           & 59.0  & 87.7  & 62.9  & 41.6  & 21.3  & 54.50 (+22.7\%) \\
        Jiuzhang3.0  & 7B Math Model       & 58.2  & 88.9  & 63.8  & 41.5  & 19.6  & 54.40 (+21.8\%) \\
        MathScale    & ChatGPT        & 58.2  & 89.5  & 63.5  & 40.3  & 20.3  & 54.36 (+22.5\%) \\
        PromptCoT    & 72B Model        & 58.2  & 87.4  & 63.5  & 39.6  & 21.3  & 54.00 (+21.6\%)\\
        ScaleQuest   & 7B Math Model       & 59.0  & 88.9  & 63.8  & 41.2  & 21.8  & 54.94 (+23.7\%) \\
        \midrule
        \rowcolor{lightyellow} RV-Syn         &       7B / 72B model         & \textbf{59.8}  & 88.8  & 65.0  & \textbf{42.2}  & \textbf{22.4}  & \textbf{55.64 (+25.3\%)} \\
        \bottomrule
    \end{tabular}
    }
    \caption{After training on 50k data, the evaluation results of our method compared with various synthesis approaches across five benchmarks.  The best ones among LLMs with the same backbone model are marked in bold.}
    \label{tab:main}
\end{table*}
\section{Experiments}
\subsection{Experimental Setup}

\paragraph{Compared Baselines.}
We compare our method with previous problem synthesis methods with publicly available datasets, including:
(1) MetaMath~\cite{metamath} introduces several question bootstrapping techniques;  
(2) Orca-Math~\cite{orcamath} augments existing datasets using an Agent-Instruct method;  
(3) MathScale~\cite{mathscale} uses topic and knowledge-point graphs to prompt new problem synthesis;  
(4) Mammoth2~\cite{mammoth2} extracts QA pairs from webpages;  
(5) Jiuzhang3.0~\cite{jiuzhang} uses math-related seed data to synthesize new QA pairs; 
(6) PromptCoT~\cite{promptcot} generate complex problems with intermediate thought; 
(7) ScaleQuest~\cite{scalequest} trains a problem generation model similar to Magpie~\cite{magpie}.

\paragraph{Tuning Data.}
Directly comparing the performance of models from previous works introduces potential unfair factors, as newer models may leverage more advanced models for backbone and answer generation (see Appendix ~\ref{Annotation} for details). 
To ensure a fair comparison, we use the same model, Qwen-2.5-Math-7B-Instruct~\cite{Yang2024qwen2.5math}
to generate answers for problems synthesized by each method and train the same backbone model. By controlling for other variables, we can focus on comparing the quality of the synthesized problems.

\paragraph{Evaluation and Metrics.}
We assess the models' performance on MATH-500~\cite{dan2021math,lightman2024step-verify}, GSM8K~\cite{gsm8k}, GSM-Hard~\cite{GaoMZ00YCN23pal}, College Math~\cite{mathscale} and OlympiadBench~\cite{HeLBHTSHHHZLQL024olympiad}.
The generated outputs are all in the form of natural language Chain-of-Thought (CoT)~\cite{wei2023chain} through greedy decoding, and we report zero-shot pass@1 accuracy.

\paragraph{Implementation Details.}
We extract seed data from the high-quality Numina-Math dataset~\cite{numina_math_datasets} to collect function nodes and fine-tuning data. The problem decomposition is performed using Qwen2.5-72B-Instruct.
We then use the fine-tuned Qwen2.5-7B-Instruct to generate computational graphs with the strategy described in \ref{ft}, followed by problem back-translation using Qwen2.5-72B-Instruct.

\subsection{Main Results}
As shown in Table \ref{tab:main}, methods focusing on synthesizing diverse new problems (\eg Mammoth2, Jiuzhang3.0, ScaleQuest) outperform approaches that augment existing data (\eg MetaMath, Orca-Math). This performance gap may be attributed to augmentation-based methods leading to new instructions that are too similar to the original ones, emphasizing the crucial role of data diversity.
Our approach outperforms traditional data synthesis methods without relying on proprietary models like GPT-4. For example, on LLaMA-3-8B-Instruct, our method outperforms the previous state-of-the-art by 6.3\%. Notably, this advantage is particularly evident on more challenging or computation-intensive datasets such as MATH-500, GSM-Hard, and OlympiadBench, further emphasizing the benefits of rational synthesis.
We present experiments on distilling Long-CoT reasoning in Appendix \ref{longcot}.

\subsection{Scalability Study}
\begin{figure}[t]
    \centering    \includegraphics[width=0.5\textwidth]{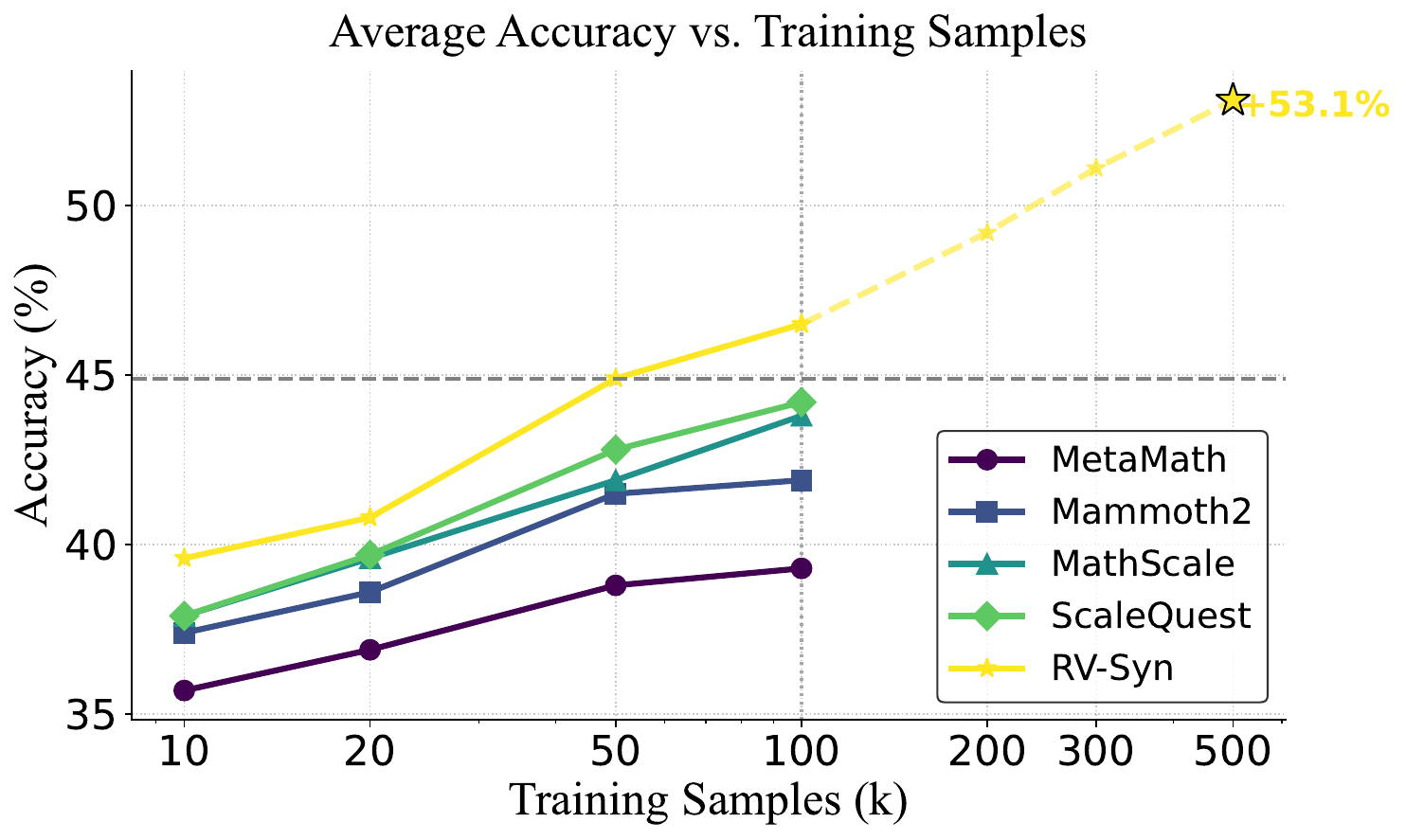}
  \caption{The performance of the model as the training data size increases. The ``Average'' refers to the mean performance across five datasets: MATH-500, GSM8K, GSM-Hard, College Math, and OlympiadBench.}
  \label{fig:scale}
\end{figure}
In this section, we study the scalability of our method.
We compare our method with previous approaches by training on scaling datasets up to 300k using LLaMA-3-8B-Instruct~\cite{llama3}.
The results in Figure \ref{fig:scale} demonstrate a strong scaling curve, with performance consistently improving as the dataset size increases. Our method consistently outperforms other methods, and notably, achieves better performance with only half the dataset size (50k vs. 100k). This advantage is particularly pronounced on challenging or computation-intensive datasets, demonstrating the superiority of our method for enhancing advanced mathematical reasoning capabilities.




\subsection{Comparison with Human-Crafted Data}
\begin{figure}[t]
    \centering    \includegraphics[width=0.5\textwidth]{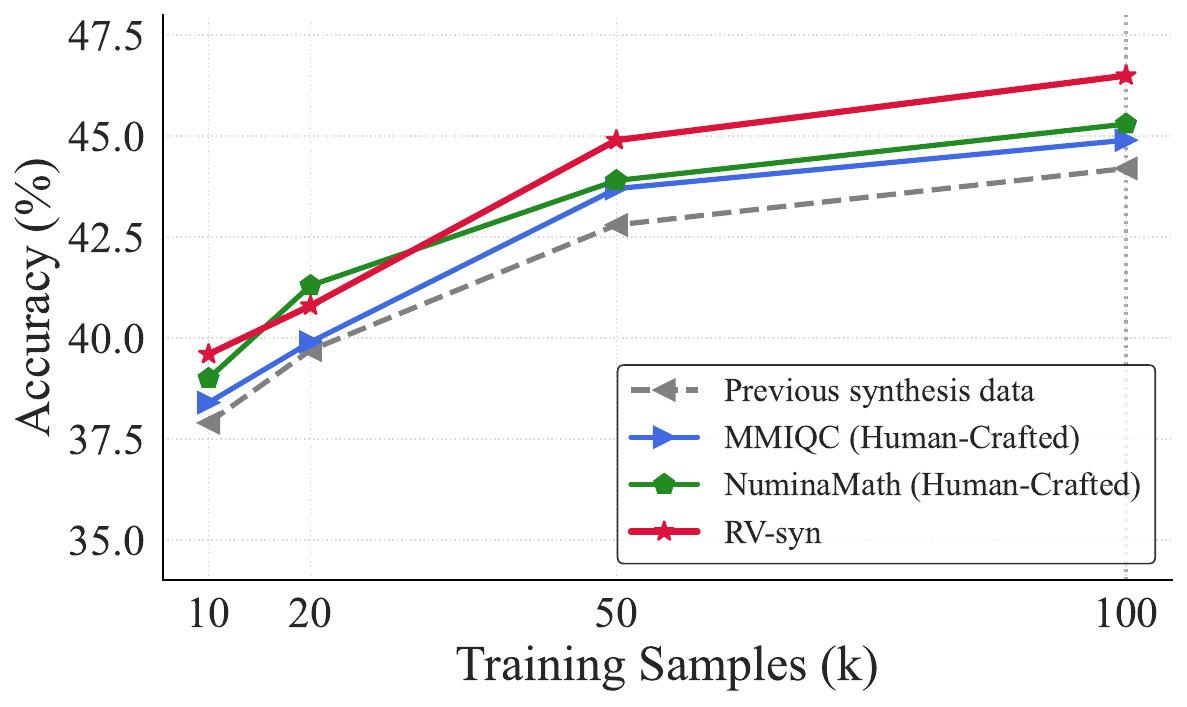}
  \caption{Comparison between RV-Syn and human-crafted data. Numina-Math serves as the seed data for our synthetic dataset. Previous state-of-the-art synthetic data methods are indicated by gray lines.}
  \label{fig:human}
\end{figure}

To rigorously benchmark the quality of our synthesized data, we compare RV-Syn against datasets that contain a significant portion of high-quality, human-crafted problems. We include two such strong baselines in our analysis: 
MMIQC~\cite{mmiqc}, which contains QA pairs from the Mathematics Stack Exchange, and
Numina-Math~\cite{numina_math_datasets}, a large collection of both real-world human-crafted and synthesized math problems, which also serves as the source of our initial seed data.
As illustrated in Figure~\ref{fig:human}, human-crafted datasets such as MMIQC and Numina-Math consistently outperform most synthetic methods, reflecting the superior quality of human-authored data over synthetic data, though they face significant scalability challenges. In contrast, RV-Syn demonstrates a clear competitive advantage: at nearly every data scale, it achieves the highest average performance, even surpassing Numina-Math, the very dataset used as its seed. This result validates that our synthesis process does not merely replicate existing examples, but instead generates novel, high-quality instructions that are more effective for model training than the original seed data itself. Consequently, RV-Syn successfully bridges the gap between scalability and quality, producing synthetic data that rivals human expertise without sacrificing performance. We provide further discussion on data diversity in Appendix \ref{diversity}.

\subsection{Analysis of Data Efficiency}
\paragraph{Higher complexity through Rational Synthesis.}
First, we posit that the solution-first, rational synthesis approach leads to higher-quality and more complex problems. A higher-quality dataset can provide richer supervision signals per example, thus improving training efficiency. To quantify this, we analyze the difficulty and efficacy of different datasets using three key metrics, following PromptCoT~\cite{promptcot}.
First, accuracy serves as an indicator of problem difficulty. Lower performance  signifies a higher level of difficulty. For datasets without ground-truth labels, we adopt the approach detailed in PromptCoT~\cite{promptcot} by leveraging a more powerful model, Qwen2.5-Math-72B-Instruct, to generate reference answers, applying self-consistency with 8 rollouts to enhance reliability.
Second, the average number of reasoning tokens provides insights into the complexity of the reasoning process. A larger average number of tokens suggests that the model necessitates more extensive reasoning steps.
Finally, the average accuracy on the benchmark after tuning reflects the dataset's contribution to improving model performance. Larger improvements in benchmark accuracy observed after fine-tuning suggest that the respective dataset is more effective in enhancing the model's capabilities.
Experimental results in Table~\ref{tab:complexity} show a strong consistency among these three metrics, and our method achieves the best results across all three difficulty evaluation metrics. Specifically, problems synthesized by RV-Syn are the most difficult (lowest accuracy at 55.6\%), require the longest reasoning chains (highest token count), and yield the largest performance improvement after fine-tuning (+34.1\%). This demonstrates that our solution-first, rational synthesis approach successfully generates more challenging and effective training instances compared to other methods.

\begin{table}[htbp]
\centering
\resizebox{0.5\textwidth}{!}{
\begin{tabular}{@{}lccccc@{}}
\toprule
\textbf{Dataset}  & \textbf{Accuracy ($\downarrow$)} & \makecell{\textbf{Avg. Reasoning} \\ \textbf{Tokens ($\uparrow$)} } & \makecell{\textbf{Avg. Benchmark} \\ \textbf{Accuracy ($\uparrow$)} }  \\
\midrule
MetaMath & 97.2 & 1,339 & 38.82 (+15.9\%) \\
Orca-Math & 86.3 & 919 & 37.92 (+13.2\%) \\
MathScale & 82.0 & 2,565 & 41.86 (+25.0\%) \\
Mammoth2 & 70.6 & 3,272 &41.54 (+24.0\%) \\
ScaleQuest & 75.8 & 2,591 &42.80 (+27.8\%) \\
Numina-Math & 72.9 & 3,858 &43.90 (+31.0\%) \\
RV-Syn & \textbf{55.6} & \textbf{4,431} &\textbf{44.92 (+34.1\%)} \\
\bottomrule
\end{tabular}
}
\caption{
Difficulty and efficacy evaluation for different datasets. \textbf{Accuracy:} Performance of Qwen2.5-Math-7B-Instruct on the problems in different datasets. \textbf{Avg. Reasoning Tokens:} Average number of tokens in reasoning processes generated by DeepSeek-R1-Distill-Qwen-7B when processing the problems. \textbf{Avg. Benchmark Accuracy:} Performance of LLaMA-3-8B-Instruct after fine-tuning on different datasets.}
\label{tab:complexity}
\end{table}

\paragraph{Higher Quality through Verifiability.}
In this part, we investigate the verifiable characteristics of our method to ensure data correctness. We sample 1,000 data points from the training data of each method, evaluate them and analyze the error rates for both problems and solutions (see Appendix ~\ref{appendix:verification validation} for details).
As shown in Table \ref{tab:correct}, firstly, the problem error rate reflects the quality of the synthesized problems. It can be observed that problems generated by humans exhibit a lower error rate, while directly synthesized problems contain more errors. Our method approaches the level of control that humans exert in problem design.
Secondly, the solution error rate reflects the extent to which each method controls the correctness of training data. Since other methods cannot naturally obtain the ground truth, their solution error rates are relatively high. Further quality control requires heavy reliance on powerful models for post-filtering, annotation, voting, or scoring, which incurs significant additional costs. In contrast, our method inherently provides the ground truth and enables rule-based matching, ensuring an extremely low error rate.
We then manually inspect our data labeled as erroneous and find that most solutions have correct final answers, with only minor errors in intermediate steps.

\begin{table}[htbp]
    \centering
    \resizebox{0.45\textwidth}{!}{
    \begin{tabular}{lccccccc}
        \toprule
        \textbf{Method}  & \makecell{\textbf{Problem Error} \\ \textbf{Rate~(\%)} \textdownarrow} & \makecell{\textbf{Solution Error} \\ \textbf{Rate~(\%)} \textdownarrow}  \\
        \midrule
        
        Mammoth2       & \underline{0.9}  & 10.6   \\
        Jiuzhang3.0   & 2.4  & 6.5   \\
        MathScale      & 1.6  & 4.9   \\
        ScaleQuest     & 5.6  & 8.0   \\
        Numina-Math        & \textbf{0.6}  & 5.5   \\
        \midrule
        RV-Syn           & \underline{0.9}  & \textbf{1.4}   \\
        \bottomrule
    \end{tabular}
    }
    \caption{Error rate analysis across different methods. The problem error refers to issues such as inconsistency or conflict in problem conditions and statements; the solution error indicates computational mistakes or misinterpretation of the problem in the solution.}
    \label{tab:correct}
\end{table}

\subsection{Ablation Study}
To provide a comprehensive understanding of RV-Syn, we conduct ablation studies to analyze the impact of our graph sampling strategies and the controllability of problem difficulty.
\paragraph{Impact of sampling strategies.}

We introduce three strategies to sample function nodes for computational graph regeneration: co-occurrence-based, topic-based, and edgeless. To evaluate their individual contributions, we synthesize 20k training samples using each strategy exclusively and train LLaMA-3-8B-Instruct.
As shown in Table \ref{tab:ablation_strategy}, each strategy excels in different areas. For example, the co-occurrence strategy is strongest on GSM8K. The edgeless strategy performs best on MATH-500. The topic-based strategy shows particular strength on the more complex GSM-Hard and OlympiadBench datasets. This validates our decision to use a mix of all three, creating a balanced and robust dataset that leverages their complementary advantages.
\begin{table}[t]
\centering
\small
\setlength{\tabcolsep}{3.5pt}
\begin{tabular}{lccccc}
\toprule
\textbf{Strategy} & \textbf{GSM8K} & \textbf{MATH} & \textbf{Hard} & \textbf{College} & \textbf{Olym.} \\
\midrule
Co-occur. & \textbf{77.9} & 35.2 & 35.3 & 26.5 & 11.0 \\
Topic & 77.2 & 35.2 & \textbf{39.2} & 26.7 & \textbf{12.9} \\
Edgeless & 77.3 & \textbf{39.6} & 37.8 & \textbf{27.1} & 12.6 \\
\bottomrule
\end{tabular}
\caption{Ablation study on different sampling strategies.}
\label{tab:ablation_strategy}
\end{table}

\paragraph{Controllability of problem difficulty.}
A key advantage of RV-Syn is the fine-grained control over synthesized problem complexity. We analyze our ability to control problem complexity by varying the number of function nodes sampled during graph regeneration. We train models on 50k datasets synthesized with a fixed number of nodes (1, 2, and 3 nodes) and evaluate them on MATH-500 and GSM-Hard.
As shown in Table~\ref{tab:ablation_nodes}, models trained on single-node problems achieve the highest performance on MATH-500, a dataset emphasizing direct application of middle-school-level concepts. In contrast, performance on the more calculation-intensive GSM-Hard improves with increasing node count, indicating that multi-node graphs effectively model complex, multi-step reasoning. This demonstrates RV-Syn’s ability to deliberately calibrate difficulty for targeted training needs.
\begin{table}[h]
\centering
\small
\begin{tabular}{lcc}
\toprule
\textbf{Sampled Nodes} & \textbf{MATH-500} & \textbf{GSM-Hard} \\
\midrule
1 Node & \textbf{49.0} & 42.6 \\
2 Nodes & 48.6 & 43.8 \\
3 Nodes & 47.2 & \textbf{44.4} \\
\bottomrule
\end{tabular}
\caption{Ablation on problem difficulty control by varying sampled node counts.}
\label{tab:ablation_nodes}
\end{table}

\subsection{Analysis of Long-CoT Distillation}
\label{longcot}
\begin{figure}[htbp]
    \centering    
    \includegraphics[width=0.48\textwidth]{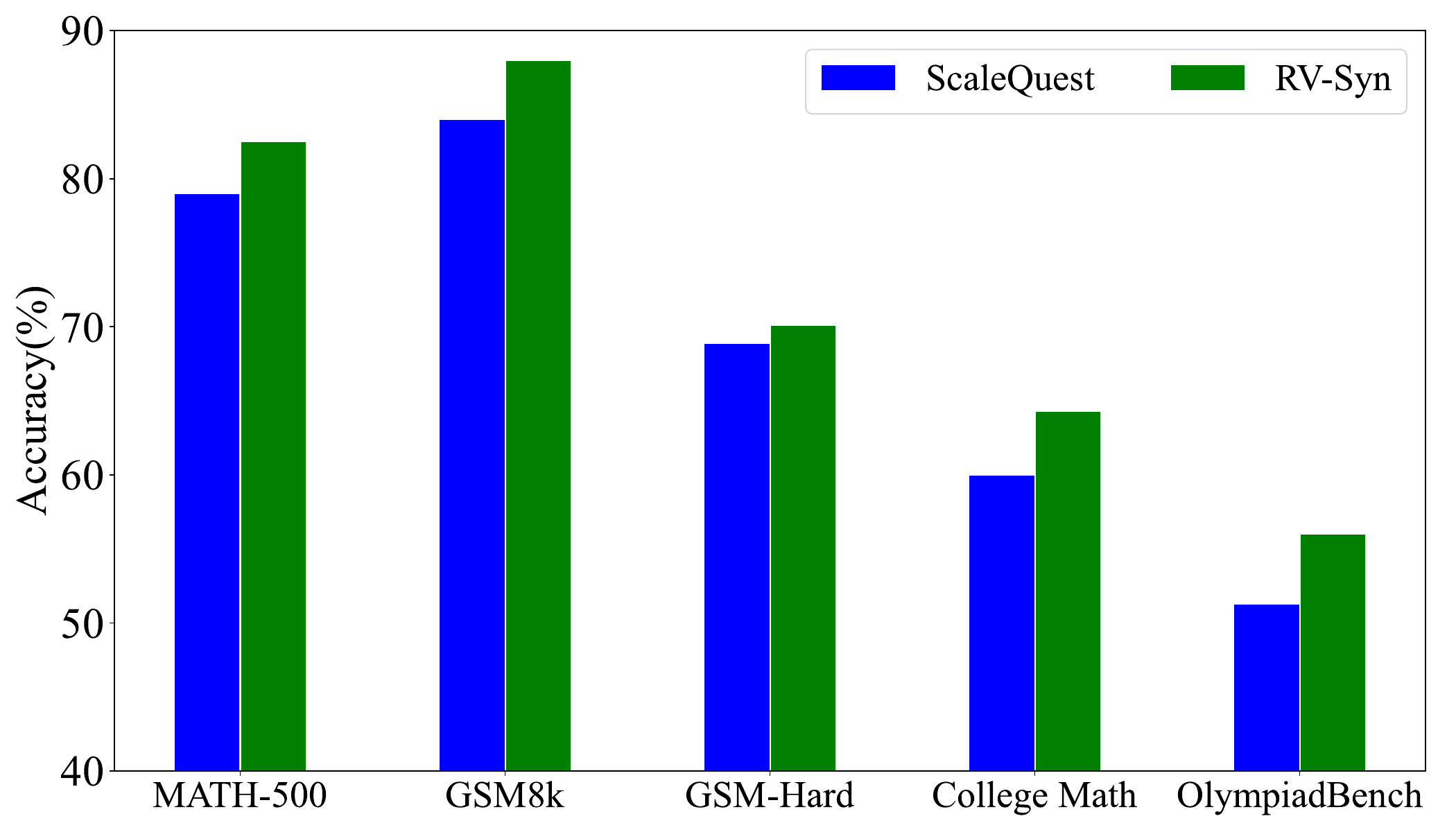}
    \caption{Performance comparison of distilling Long-CoT on 20k data.}
    \label{fig:longcot}
\end{figure}
In addition to short-CoT, we also explore the ability of our data to distill Long-CoT abilities. We replace the answer generation model with DeepSeek-R1-Distill-Qwen-7B~\cite{r1} and train Qwen2.5-7B-Instruct on a 20k dataset, comparing our method with ScaleQuest. During the evaluation, we set the maximum sequence length to 16k and exclude samples where both models exceed the maximal length. We show the results in Figure \ref{fig:longcot}. We can see that our synthesis method not only performs well on short-CoT but also excels in distilling Long-CoT compared to the baseline.

\section{Conclusion}
In this paper, we introduced RV-Syn, a novel approach to mathematical problem synthesis that enhances both rationale and verifiability. It first constructed a graph-format function library, which was composed of a large number of mathematical functions and was scalable to insert new ones. Then, it generated the new solutions by combining functions into computation graphs in a controllable manner, which can be then executed to support solution verification. Finally, it back-translated these computation graphs to problems. Our experiments demonstrate that RV-Syn outperforms existing methods, achieving higher performance with less data. This efficiency highlights its potential for improving the reasoning capabilities of LLMs while minimizing resource consumption.
RV-Syn represents a significant advancement in reasoning data generation, offering a scalable and reliable framework. Future work can explore extending it to other reasoning domains and further refining control mechanisms for broader applicability.

\section*{Limitations}
Despite the promising results demonstrated by the RV-Syn method, several limitations remain. First, the approach relies heavily on the quality and diversity of the initial function library. While we have curated a comprehensive set of functions from a large scale of existing synthetic datasets, there may still be gaps in representing certain mathematical concepts, potentially limiting the scope of problems that can be synthesized.
Second, although our experiments show improved performance across several benchmarks, the generalization of RV-Syn to other domains beyond mathematics remains unexplored, such as the code domain. Future work should investigate the adaptability of this approach to different types of reasoning tasks and datasets.
We aim to address these limitations in future research by expanding the function library, optimizing computational efficiency, and exploring cross-domain applications.

\section*{Acknowledgments}
This paper was partially supported by the National Natural Science Foundation of China No. 92470205 and Beijing Major Science and Technology Project under Contract no. Z251100008425002. Xin Zhao is the corresponding author.

\bibliography{custom}

@inproceedings{metamath,
  author       = {Longhui Yu and
                  Weisen Jiang and
                  Han Shi and
                  Jincheng Yu and
                  Zhengying Liu and
                  Yu Zhang and
                  James T. Kwok and
                  Zhenguo Li and
                  Adrian Weller and
                  Weiyang Liu},
  title        = {MetaMath: Bootstrap Your Own Mathematical Questions for Large Language
                  Models},
  booktitle    = {{ICLR}},
  publisher    = {OpenReview.net},
  year         = {2024}
}

@article{orcamath,
  author       = {Arindam Mitra and
                  Hamed Khanpour and
                  Corby Rosset and
                  Ahmed Awadallah},
  title        = {Orca-Math: Unlocking the potential of SLMs in Grade School Math},
  journal      = {CoRR},
  volume       = {abs/2402.14830},
  year         = {2024}
}

@article{mammoth2,
  author       = {Xiang Yue and
                  Tuney Zheng and
                  Ge Zhang and
                  Wenhu Chen},
  title        = {MAmmoTH2: Scaling Instructions from the Web},
  journal      = {CoRR},
  volume       = {abs/2405.03548},
  year         = {2024}
}

@article{jiuzhang,
  author       = {Kun Zhou and
                  Beichen Zhang and
                  Jiapeng Wang and
                  Zhipeng Chen and
                  Wayne Xin Zhao and
                  Jing Sha and
                  Zhichao Sheng and
                  Shijin Wang and
                  Ji-Rong Wen},
  title        = {JiuZhang3.0: Efficiently Improving Mathematical Reasoning by Training
                  Small Data Synthesis Models},
  journal      = {CoRR},
  volume       = {abs/2405.14365},
  year         = {2024}
}

@inproceedings{mathscale,
  author       = {Zhengyang Tang and
                  Xingxing Zhang and
                  Benyou Wang and
                  Furu Wei},
  title        = {MathScale: Scaling Instruction Tuning for Mathematical Reasoning},
  booktitle    = {{ICML}},
  publisher    = {OpenReview.net},
  year         = {2024}
}

@article{mmiqc,
  author       = {Haoxiong Liu and
                  Yifan Zhang and
                  Yifan Luo and
                  Andrew Chi{-}Chih Yao},
  title        = {Augmenting Math Word Problems via Iterative Question Composing},
  journal      = {CoRR},
  volume       = {abs/2401.09003},
  year         = {2024}
}

@article{scalequest,
  author       = {Yuyang Ding and
                  Xinyu Shi and
                  Xiaobo Liang and
                  Juntao Li and
                  Qiaoming Zhu and
                  Min Zhang},
  title        = {Unleashing Reasoning Capability of LLMs via Scalable Question Synthesis
                  from Scratch},
  journal      = {CoRR},
  volume       = {abs/2410.18693},
  year         = {2024}
}

@misc{numina_math_datasets,
  author = {Jia LI and Edward Beeching and Lewis Tunstall and Ben Lipkin and Roman Soletskyi and Shengyi Costa Huang and Kashif Rasul and Longhui Yu and Albert Jiang and Ziju Shen and Zihan Qin and Bin Dong and Li Zhou and Yann Fleureau and Guillaume Lample and Stanislas Polu},
  title = {NuminaMath},
  year = {2024},
  publisher = {Numina},
  journal = {Hugging Face repository},
  howpublished = {\url{https://huggingface.co/AI-MO/NuminaMath-CoT}}
}

@article{Yang2024qwen2.5math,
  author       = {An Yang and
                  Beichen Zhang and
                  Binyuan Hui and
                  Bofei Gao and
                  Bowen Yu and
                  Chengpeng Li and
                  Dayiheng Liu and
                  Jianhong Tu and
                  Jingren Zhou and
                  Junyang Lin and
                  Keming Lu and
                  Mingfeng Xue and
                  Runji Lin and
                  Tianyu Liu and
                  Xingzhang Ren and
                  Zhenru Zhang},
  title        = {Qwen2.5-Math Technical Report: Toward Mathematical Expert Model via
                  Self-Improvement},
  journal      = {CoRR},
  volume       = {abs/2409.12122},
  year         = {2024}
}

@misc{llama3,
  title={Llama 3 Model Card},
  author={AI@Meta},
  year={2024},
  url = {https://github.com/meta-llama/llama3/blob/main/MODEL_CARD.md}
}

@inproceedings{dan2021math,
  author       = {Dan Hendrycks and
                  Collin Burns and
                  Saurav Kadavath and
                  Akul Arora and
                  Steven Basart and
                  Eric Tang and
                  Dawn Song and
                  Jacob Steinhardt},
  editor       = {Joaquin Vanschoren and
                  Sai{-}Kit Yeung},
  title        = {Measuring Mathematical Problem Solving With the {MATH} Dataset},
  booktitle    = {Proceedings of the Neural Information Processing Systems Track on
                  Datasets and Benchmarks 1, NeurIPS Datasets and Benchmarks 2021, December
                  2021, virtual},
  year         = {2021},
  url          = {https://datasets-benchmarks-proceedings.neurips.cc/paper/2021/hash/be83ab3ecd0db773eb2dc1b0a17836a1-Abstract-round2.html},
  bibsource    = {dblp computer science bibliography, https://dblp.org}
}

@inproceedings{HeLBHTSHHHZLQL024olympiad,
  author       = {Chaoqun He and
                  Renjie Luo and
                  Yuzhuo Bai and
                  Shengding Hu and
                  Zhen Leng Thai and
                  Junhao Shen and
                  Jinyi Hu and
                  Xu Han and
                  Yujie Huang and
                  Yuxiang Zhang and
                  Jie Liu and
                  Lei Qi and
                  Zhiyuan Liu and
                  Maosong Sun},
  title        = {OlympiadBench: {A} Challenging Benchmark for Promoting {AGI} with
                  Olympiad-Level Bilingual Multimodal Scientific Problems},
  booktitle    = {{ACL} {(1)}},
  pages        = {3828--3850},
  publisher    = {Association for Computational Linguistics},
  year         = {2024}
}

@inproceedings{GaoMZ00YCN23pal,
  author       = {Luyu Gao and
                  Aman Madaan and
                  Shuyan Zhou and
                  Uri Alon and
                  Pengfei Liu and
                  Yiming Yang and
                  Jamie Callan and
                  Graham Neubig},
  title        = {{PAL:} Program-aided Language Models},
  booktitle    = {{ICML}},
  series       = {Proceedings of Machine Learning Research},
  volume       = {202},
  pages        = {10764--10799},
  publisher    = {{PMLR}},
  year         = {2023}
}

@inproceedings{lightman2024step-verify,
  author       = {Hunter Lightman and
                  Vineet Kosaraju and
                  Yuri Burda and
                  Harrison Edwards and
                  Bowen Baker and
                  Teddy Lee and
                  Jan Leike and
                  John Schulman and
                  Ilya Sutskever and
                  Karl Cobbe},
  title        = {Let's Verify Step by Step},
  booktitle    = {The Twelfth International Conference on Learning Representations,
                  {ICLR} 2024, Vienna, Austria, May 7-11, 2024},
  publisher    = {OpenReview.net},
  year         = {2024}
}

@inproceedings{wei2023chain,
  author       = {Jason Wei and
                  Xuezhi Wang and
                  Dale Schuurmans and
                  Maarten Bosma and
                  Brian Ichter and
                  Fei Xia and
                  Ed H. Chi and
                  Quoc V. Le and
                  Denny Zhou},
  title        = {Chain-of-Thought Prompting Elicits Reasoning in Large Language Models},
  booktitle    = {NeurIPS},
  year         = {2022}
}

@article{syne,
  author       = {Jie Chen and
                  Zhipeng Chen and
                  Jiapeng Wang and
                  Kun Zhou and
                  Yutao Zhu and
                  Jinhao Jiang and
                  Yingqian Min and
                  Wayne Xin Zhao and
                  Zhicheng Dou and
                  Jiaxin Mao and
                  Yankai Lin and
                  Ruihua Song and
                  Jun Xu and
                  Xu Chen and
                  Rui Yan and
                  Zhewei Wei and
                  Di Hu and
                  Wenbing Huang and
                  Ji-Rong Wen},
  title        = {Towards Effective and Efficient Continual Pre-training of Large Language
                  Models},
  journal      = {CoRR},
  volume       = {abs/2407.18743},
  year         = {2024}
}

@article{gsm8k,
  author       = {Karl Cobbe and
                  Vineet Kosaraju and
                  Mohammad Bavarian and
                  Mark Chen and
                  Heewoo Jun and
                  Lukasz Kaiser and
                  Matthias Plappert and
                  Jerry Tworek and
                  Jacob Hilton and
                  Reiichiro Nakano and
                  Christopher Hesse and
                  John Schulman},
  title        = {Training Verifiers to Solve Math Word Problems},
  journal      = {CoRR},
  volume       = {abs/2110.14168},
  year         = {2021}
}

@article{magpie,
  author       = {Zhangchen Xu and
                  Fengqing Jiang and
                  Luyao Niu and
                  Yuntian Deng and
                  Radha Poovendran and
                  Yejin Choi and
                  Bill Yuchen Lin},
  title        = {Magpie: Alignment Data Synthesis from Scratch by Prompting Aligned
                  LLMs with Nothing},
  journal      = {CoRR},
  volume       = {abs/2406.08464},
  year         = {2024}
}

@article{wizardmath,
  author       = {Haipeng Luo and
                  Qingfeng Sun and
                  Can Xu and
                  Pu Zhao and
                  Jianguang Lou and
                  Chongyang Tao and
                  Xiubo Geng and
                  Qingwei Lin and
                  Shifeng Chen and
                  Dongmei Zhang},
  title        = {WizardMath: Empowering Mathematical Reasoning for Large Language Models
                  via Reinforced Evol-Instruct},
  journal      = {CoRR},
  volume       = {abs/2308.09583},
  year         = {2023}
}

@inproceedings{mugglemath,
  author       = {Chengpeng Li and
                  Zheng Yuan and
                  Hongyi Yuan and
                  Guanting Dong and
                  Keming Lu and
                  Jiancan Wu and
                  Chuanqi Tan and
                  Xiang Wang and
                  Chang Zhou},
  title        = {MuggleMath: Assessing the Impact of Query and Response Augmentation
                  on Math Reasoning},
  booktitle    = {{ACL} {(1)}},
  pages        = {10230--10258},
  publisher    = {Association for Computational Linguistics},
  year         = {2024}
}

@inproceedings{mathgenie,
  author       = {Zimu Lu and
                  Aojun Zhou and
                  Houxing Ren and
                  Ke Wang and
                  Weikang Shi and
                  Junting Pan and
                  Mingjie Zhan and
                  Hongsheng Li},
  title        = {MathGenie: Generating Synthetic Data with Question Back-translation
                  for Enhancing Mathematical Reasoning of LLMs},
  booktitle    = {{ACL} {(1)}},
  pages        = {2732--2747},
  publisher    = {Association for Computational Linguistics},
  year         = {2024}
}

@article{kpmath,
  author       = {Yiming Huang and
                  Xiao Liu and
                  Yeyun Gong and
                  Zhibin Gou and
                  Yelong Shen and
                  Nan Duan and
                  Weizhu Chen},
  title        = {Key-Point-Driven Data Synthesis with its Enhancement on Mathematical
                  Reasoning},
  journal      = {CoRR},
  volume       = {abs/2403.02333},
  year         = {2024}
}

@inproceedings{tora,
  author       = {Zhibin Gou and
                  Zhihong Shao and
                  Yeyun Gong and
                  Yelong Shen and
                  Yujiu Yang and
                  Minlie Huang and
                  Nan Duan and
                  Weizhu Chen},
  title        = {ToRA: {A} Tool-Integrated Reasoning Agent for Mathematical Problem
                  Solving},
  booktitle    = {{ICLR}},
  publisher    = {OpenReview.net},
  year         = {2024}
}

@inproceedings{mathcoder,
  author       = {Ke Wang and
                  Houxing Ren and
                  Aojun Zhou and
                  Zimu Lu and
                  Sichun Luo and
                  Weikang Shi and
                  Renrui Zhang and
                  Linqi Song and
                  Mingjie Zhan and
                  Hongsheng Li},
  title        = {MathCoder: Seamless Code Integration in LLMs for Enhanced Mathematical
                  Reasoning},
  booktitle    = {{ICLR}},
  publisher    = {OpenReview.net},
  year         = {2024}
}

@article{Shao2024DeepSeekMath,
  author       = {Zhihong Shao and
                  Peiyi Wang and
                  Qihao Zhu and
                  Runxin Xu and
                  Junxiao Song and
                  Mingchuan Zhang and
                  Y. K. Li and
                  Y. Wu and
                  Daya Guo},
  title        = {DeepSeekMath: Pushing the Limits of Mathematical Reasoning in Open
                  Language Models},
  journal      = {CoRR},
  volume       = {abs/2402.03300},
  year         = {2024}
}

@inproceedings{DBLP:conf/emnlp/ChenWCS024,
  author       = {Nuo Chen and
                  Ning Wu and
                  Jianhui Chang and
                  Linjun Shou and
                  Jia Li},
  title        = {ControlMath: Controllable Data Generation Promotes Math Generalist
                  Models},
  booktitle    = {{EMNLP}},
  pages        = {12201--12217},
  publisher    = {Association for Computational Linguistics},
  year         = {2024}
}

@article{pot,
  author       = {Wenhu Chen and
                  Xueguang Ma and
                  Xinyi Wang and
                  William W. Cohen},
  title        = {Program of Thoughts Prompting: Disentangling Computation from Reasoning
                  for Numerical Reasoning Tasks},
  journal      = {Trans. Mach. Learn. Res.},
  volume       = {2023},
  year         = {2023}
}

@inproceedings{react,
  author       = {Shunyu Yao and
                  Jeffrey Zhao and
                  Dian Yu and
                  Nan Du and
                  Izhak Shafran and
                  Karthik R. Narasimhan and
                  Yuan Cao},
  title        = {ReAct: Synergizing Reasoning and Acting in Language Models},
  booktitle    = {{ICLR}},
  publisher    = {OpenReview.net},
  year         = {2023}
}

@inproceedings{reflexion,
  author       = {Noah Shinn and
                  Federico Cassano and
                  Ashwin Gopinath and
                  Karthik Narasimhan and
                  Shunyu Yao},
  title        = {Reflexion: language agents with verbal reinforcement learning},
  booktitle    = {NeurIPS},
  year         = {2023}
}

@inproceedings{llemma,
  author       = {Zhangir Azerbayev and
                  Hailey Schoelkopf and
                  Keiran Paster and
                  Marco Dos Santos and
                  Stephen Marcus McAleer and
                  Albert Q. Jiang and
                  Jia Deng and
                  Stella Biderman and
                  Sean Welleck},
  title        = {Llemma: An Open Language Model for Mathematics},
  booktitle    = {{ICLR}},
  publisher    = {OpenReview.net},
  year         = {2024}
}

@article{r1,
  title={Deepseek-r1: Incentivizing reasoning capability in llms via reinforcement learning},
  author={Guo, Daya and Yang, Dejian and Zhang, Haowei and Song, Junxiao and Zhang, Ruoyu and Xu, Runxin and Zhu, Qihao and Ma, Shirong and Wang, Peiyi and Bi, Xiao and others},
  journal={arXiv preprint arXiv:2501.12948},
  year={2025}
}

@misc{openai2024reasoning,
  author = {OpenAI},
  title = {Learning to Reason with LLMs},
  year = {2024},
  url = {https://openai.com/index/learning-to-reason-with-llms}
}

@article{survey,
  author       = {Wayne Xin Zhao and
                  Kun Zhou and
                  Junyi Li and
                  Tianyi Tang and
                  Xiaolei Wang and
                  Yupeng Hou and
                  Yingqian Min and
                  Beichen Zhang and
                  Junjie Zhang and
                  Zican Dong and
                  Yifan Du and
                  Chen Yang and
                  Yushuo Chen and
                  Zhipeng Chen and
                  Jinhao Jiang and
                  Ruiyang Ren and
                  Yifan Li and
                  Xinyu Tang and
                  Zikang Liu and
                  Peiyu Liu and
                  Jian{-}Yun Nie and
                  Ji{-}Rong Wen},
  title        = {A Survey of Large Language Models},
  journal      = {CoRR},
  volume       = {abs/2303.18223},
  year         = {2023}
}

@inproceedings{wizardlm,
  author       = {Can Xu and
                  Qingfeng Sun and
                  Kai Zheng and
                  Xiubo Geng and
                  Pu Zhao and
                  Jiazhan Feng and
                  Chongyang Tao and
                  Qingwei Lin and
                  Daxin Jiang},
  title        = {WizardLM: Empowering Large Pre-Trained Language Models to Follow Complex
                  Instructions},
  booktitle    = {{ICLR}},
  publisher    = {OpenReview.net},
  year         = {2024}
}

@article{scratch,
  author       = {Haoran Li and
                  Qingxiu Dong and
                  Zhengyang Tang and
                  Chaojun Wang and
                  Xingxing Zhang and
                  Haoyang Huang and
                  Shaohan Huang and
                  Xiaolong Huang and
                  Zeqiang Huang and
                  Dongdong Zhang and
                  Yuxian Gu and
                  Xin Cheng and
                  Xun Wang and
                  Si{-}Qing Chen and
                  Li Dong and
                  Wei Lu and
                  Zhifang Sui and
                  Benyou Wang and
                  Wai Lam and
                  Furu Wei},
  title        = {Synthetic Data (Almost) from Scratch: Generalized Instruction Tuning
                  for Language Models},
  journal      = {CoRR},
  volume       = {abs/2402.13064},
  year         = {2024}
}

@article{AI-Assisted,
  author       = {Vedant Shah and
                  Dingli Yu and
                  Kaifeng Lyu and
                  Simon Park and
                  Nan Rosemary Ke and
                  Michael Mozer and
                  Yoshua Bengio and
                  Sanjeev Arora and
                  Anirudh Goyal},
  title        = {AI-Assisted Generation of Difficult Math Questions},
  journal      = {CoRR},
  volume       = {abs/2407.21009},
  year         = {2024}
}

@article{xwin,
  author       = {Chen Li and
                  Weiqi Wang and
                  Jingcheng Hu and
                  Yixuan Wei and
                  Nanning Zheng and
                  Han Hu and
                  Zheng Zhang and
                  Houwen Peng},
  title        = {Common 7B Language Models Already Possess Strong Math Capabilities},
  journal      = {CoRR},
  volume       = {abs/2403.04706},
  year         = {2024}
}

@inproceedings{dart,
  author       = {Yuxuan Tong and
                  Xiwen Zhang and
                  Rui Wang and
                  Ruidong Wu and
                  Junxian He},
  title        = {DART-Math: Difficulty-Aware Rejection Tuning for Mathematical Problem-Solving},
  booktitle    = {NeurIPS},
  year         = {2024}
}

@inproceedings{MUSTARD,
  author       = {Yinya Huang and
                  Xiaohan Lin and
                  Zhengying Liu and
                  Qingxing Cao and
                  Huajian Xin and
                  Haiming Wang and
                  Zhenguo Li and
                  Linqi Song and
                  Xiaodan Liang},
  title        = {{MUSTARD:} Mastering Uniform Synthesis of Theorem and Proof Data},
  booktitle    = {{ICLR}},
  publisher    = {OpenReview.net},
  year         = {2024}
}

@inproceedings{Automatic_Instruction_Evolving,
  author       = {Weihao Zeng and
                  Can Xu and
                  Yingxiu Zhao and
                  Jian{-}Guang Lou and
                  Weizhu Chen},
  title        = {Automatic Instruction Evolving for Large Language Models},
  booktitle    = {{EMNLP}},
  pages        = {6998--7018},
  publisher    = {Association for Computational Linguistics},
  year         = {2024}
}

@article{scaling,
  author       = {Zheng Yuan and
                  Hongyi Yuan and
                  Chengpeng Li and
                  Guanting Dong and
                  Chuanqi Tan and
                  Chang Zhou},
  title        = {Scaling Relationship on Learning Mathematical Reasoning with Large
                  Language Models},
  journal      = {CoRR},
  volume       = {abs/2308.01825},
  year         = {2023}
}

@inproceedings{least_to_most,
  author       = {Denny Zhou and
                  Nathanael Sch{\"a}rli and
                  Le Hou and
                  Jason Wei and
                  Nathan Scales and
                  Xuezhi Wang and
                  Dale Schuurmans and
                  Claire Cui and
                  Olivier Bousquet and
                  Quoc V. Le and
                  Ed H. Chi},
  title        = {Least-to-Most Prompting Enables Complex Reasoning in Large Language
                  Models},
  booktitle    = {{ICLR}},
  publisher    = {OpenReview.net},
  year         = {2023}
}

@article{DBLP:journals/corr/abs-2401-05384,
  author       = {Nuo Chen and
                  Hongguang Li and
                  Baoyuan Wang and
                  Jia Li},
  title        = {From Good to Great: Improving Math Reasoning with Tool-Augmented Interleaf
                  Prompting},
  journal      = {CoRR},
  volume       = {abs/2401.05384},
  year         = {2024}
}

@inproceedings{openmathinstruct,
  author       = {Shubham Toshniwal and
                  Ivan Moshkov and
                  Sean Narenthiran and
                  Daria Gitman and
                  Fei Jia and
                  Igor Gitman},
  title        = {OpenMathInstruct-1: {A} 1.8 Million Math Instruction Tuning Dataset},
  booktitle    = {NeurIPS},
  year         = {2024}
}

@article{DBLP:journals/corr/abs-2310-05506,
  author       = {Chengpeng Li and
                  Zheng Yuan and
                  Hongyi Yuan and
                  Guanting Dong and
                  Keming Lu and
                  Jiancan Wu and
                  Chuanqi Tan and
                  Xiang Wang and
                  Chang Zhou},
  title        = {Query and Response Augmentation Cannot Help Out-of-domain Math Reasoning
                  Generalization},
  journal      = {CoRR},
  volume       = {abs/2310.05506},
  year         = {2023}
}

@article{promptcot,
  author       = {Xueliang Zhao and
                  Wei Wu and
                  Jian Guan and
                  Lingpeng Kong},
  title        = {PromptCoT: Synthesizing Olympiad-level Problems for Mathematical Reasoning
                  in Large Language Models},
  journal      = {CoRR},
  volume       = {abs/2503.02324},
  year         = {2025}
}

@article{codeplan,
  author       = {Jiaxin Wen and
                  Jian Guan and
                  Hongning Wang and
                  Wei Wu and
                  Minlie Huang},
  title        = {CodePlan: Unlocking Reasoning Potential in Large Langauge Models by
                  Scaling Code-form Planning},
  journal      = {CoRR},
  volume       = {abs/2409.12452},
  year         = {2024}
}

\appendix

\section{Prompts for Data Generation Pipeline}
\label{appendix:prompts}
In this part, we present the prompts used in the three main components of our method: Problem Decomposition, Computational Graph Regeneration, and Problem Back-translation.
\begin{promptbox}[Prompt for Problem Decomposition]{lightblue}
Your task is to solve the following math problem using Python code, employing a `main` function along with several custom operation functions.\\
\\
Problem: \{PROBLEM\}\\
\\
Answer the question step by step first, understand the logic of the problem. Then:\\

1. Represent calculation operations as functions, where each function specifies its functionality, input parameter types, and output return types.\\
2. Each operation function should represent a complete, meaningful computational skill or operation, and should be reusable across different problems.\\
3. Simple operations such as addition, subtraction, multiplication and division do not need to be expressed as functions.\\
4. Each function should have a specific, independent purpose for performing a particular calculation. You may use functions from standard or imported libraries (like SymPy, math, etc.; Import within the function itself), but **cannot** use other custom operation functions.\\
5. The operation functions should be flexible, not hardcoded, and able to handle various scenarios by accepting different parameters. \\
6. Solve the problem by calling the operation functions within the main function.\\
7. Only print the answer without any text.
\end{promptbox}

\begin{promptbox}[Prompt for Function Labelling]{lightblue}
Please analyze the following mathematical function operation, judge its functional correctness and assign it a subject category.\\
\\
\{function\}\\
\\
- First analyze the function and evaluate its correctness. Output Yes or No for correctness judgement.\\
- There are 7 possible subject categories to choose from: Prealgebra, Intermediate Algebra, Algebra, Number Theory, Precalculus, Geometry, Counting \& Probability, and Miscellaneous.\\
\\
Output example:\\
<analyse>analyze here</analyse>\\
<correctness>Yes</correctness>\\
<subject>Number Theory</subject> 
\end{promptbox}

\begin{promptbox}[Prompt for Computational Graph Regeneration Fine-tuning]{lightblue}
Given a set of mathematical operation functions:  \\
\verb|```|\\
\{FUNCTIONS\}\\
\verb|```|

Write a `main` function that intelligently combines these operations to solve a meaningful and non-trivial mathematical problem.
\end{promptbox}

\begin{promptbox}[Prompt for Problem Back-translation]{lightblue}
Given a piece of code that solves a mathematical problem:\\  
\verb|```|\\
\{CODE\}\\
\verb|```|

First, understand what the code snippet is doing within the `<analysis></analysis>` tags. Then, identify the original problem the functions are solving and output it within the `<problem></problem>` tags. Finally, rewrite the problem concisely, in a manner similar to questions found in textbooks or test papers, and output it within the `<final problem></final problem>` tags. The question should have only one answer.
\end{promptbox}

\begin{promptbox}[Prompt for Correctness Verification]{lightblue}
Your goal is to determine if the given problem and its solution contain mistakes.\\
\\
If both the problem and the solution are correct, output 'Correct'.\\
If the problem contains error such as inconsistency, conflict or no solution, output 'Problem Error'.\\
If the problem is correct but the solution contains error, output 'Solution Error'.\\
\\
Problem: \{problem\}\\
\\
Solution: \{solution\}
\end{promptbox}

\section{Details for Correctness Evaluation }\label{appendix:verification validation}
Considering the substantial cost associated with human evaluations, we use one of the most powerful models currently available, GPT-4o, to conduct correctness evaluation. To ensure its validity, we conducted an experiment using GPT-4o to evaluate the accuracy of Qwen2.5-Math-7B-Instruct on the benchmark test set and compared the results with ground-truth evaluations. As shown in Table \ref{tab:gpt4o_confusion_matrix}, the evaluation results—74.25\% (GPT-4o) vs. 76.61\% (Ground truth)—demonstrate a close alignment. False positive samples account for only 1.89\% of all cases, while the false negative rate is even higher, indicating that GPT-4o tends to be more stringent in its evaluation. This is likely because it not only verifies the final answer but also checks intermediate steps. Considering that even ground-truth-based evaluations may contain some degree of error, we believe this discrepancy falls within an acceptable range. The prompt we use is shown in Appendix \ref{appendix:prompts}.

\begin{table}[htbp]
\small
\centering
\resizebox{0.48\textwidth}{!}{
\begin{tabular}{ccccc}
\toprule
 &  & \multicolumn{2}{c}{\textbf{Ground Truth}} & \\
\cmidrule{3-4}
 &  & \textbf{Correct (+)} & \textbf{Incorrect (–)} & \textbf{Total} \\
\midrule
\multirow{2}{*}{\textbf{GPT-4o}} 
 & \textbf{Correct (+)} & 0.7236 & 0.0189 & 0.7425 \\
 & \textbf{Incorrect (–)} & 0.0425 & 0.2150 & 0.2575 \\
\midrule
\textbf{Total} &  & 0.7661 & 0.2339 & 1.0000 \\
\bottomrule
\end{tabular}
}
\caption{Confusion matrix for GPT-4o evaluation.}
\label{tab:gpt4o_confusion_matrix}
\end{table}

\section{Example of Problem Decomposition}\label{appendix:format}
\begin{figure}[ht]
    \centering    \includegraphics[width=0.5\textwidth]{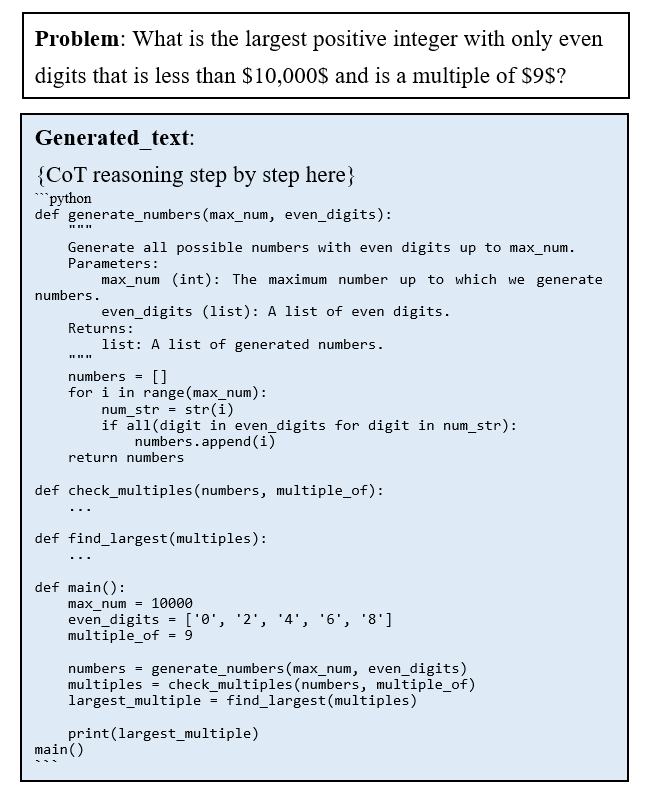}
  
  \label{fig:format}
\end{figure}

\section{Analysis of Data Diversity}
\label{diversity}
A key limitation of data augmentation methods like MetaMath~\cite{metamath} is their tendency to produce problems that are structurally similar to the original seeds. This often results from the inherent paraphrasing nature of in-context learning, leading to limited diversity. For example, a seed problem asking to ``choose 3 captains from a team of 11 people'' might be augmented to ``choose 2 captains from a team of 9 people'', a trivial variation.
RV-Syn also utilizes seed data, but in a fundamentally different manner. Instead of rewriting whole problems, RV-Syn decomposes them to extract a library of atomic mathematical operations (functions). Our synthesis process then generates new problems not by modifying an existing problem, but by creating entirely new computational graphs by combining these atomic functions. This combinatorial approach allows for the creation of novel reasoning paths and complex problem structures that can be fundamentally different from any single seed problem, thereby fostering genuine diversity.

To empirically validate this claim, we quantitatively compare the diversity of data generated by RV-Syn against MetaMath, relative to the seed data. We employed three metrics: 
$\Delta$ Mean of Embedding Variance: The change in the average variance of problem embeddings. A positive value indicates increased diversity.
Centroid Cosine Distance and Maximum Mean Discrepancy (MMD$^2$): Both measure the dissimilarity between the generated data distribution and the original seed distribution. Larger values signify greater divergence.

\begin{table}[h]
\centering
\resizebox{0.48\textwidth}{!}{%
\begin{tabular}{@{}lccc@{}}
\toprule
\textbf{Method} & \makecell{\textbf{$\Delta$ Mean of} \\ \textbf{Emb. Variance}} & \makecell{\textbf{Centroid} \\ \textbf{Cosine Dist.}} & \textbf{MMD$^2$} \\
\midrule
MetaMath & -0.0017 & 0.0170 & 0.000825 \\
\textbf{RV-Syn} & \textbf{+0.0019} & \textbf{0.0398} & \textbf{0.002493} \\
\bottomrule
\end{tabular}%
}
\caption{Diversity and similarity analysis. RV-Syn increases embedding variance and generates data that is more distinct from the seed set compared to MetaMath.}
\label{tab:diversity}
\end{table}

The results, presented in Table~\ref{tab:diversity}, strongly support our hypothesis. MetaMath shows a negative change in embedding variance, suggesting it slightly reduces diversity. In sharp contrast, RV-Syn increases diversity. Furthermore, the significantly larger Centroid Cosine Distance and MMD$^2$ values for RV-Syn confirm that its output diverges more substantially from the seed distribution. This demonstrates that our function-centric, combinatorial synthesis is more effective at producing novel and diverse problems than simple seed-based augmentation.

\section{Statistics of the Synthesis Pipeline}\label{app:statistic}
\begin{figure*}[htbp]
    \centering    
    \includegraphics[width=1\textwidth]{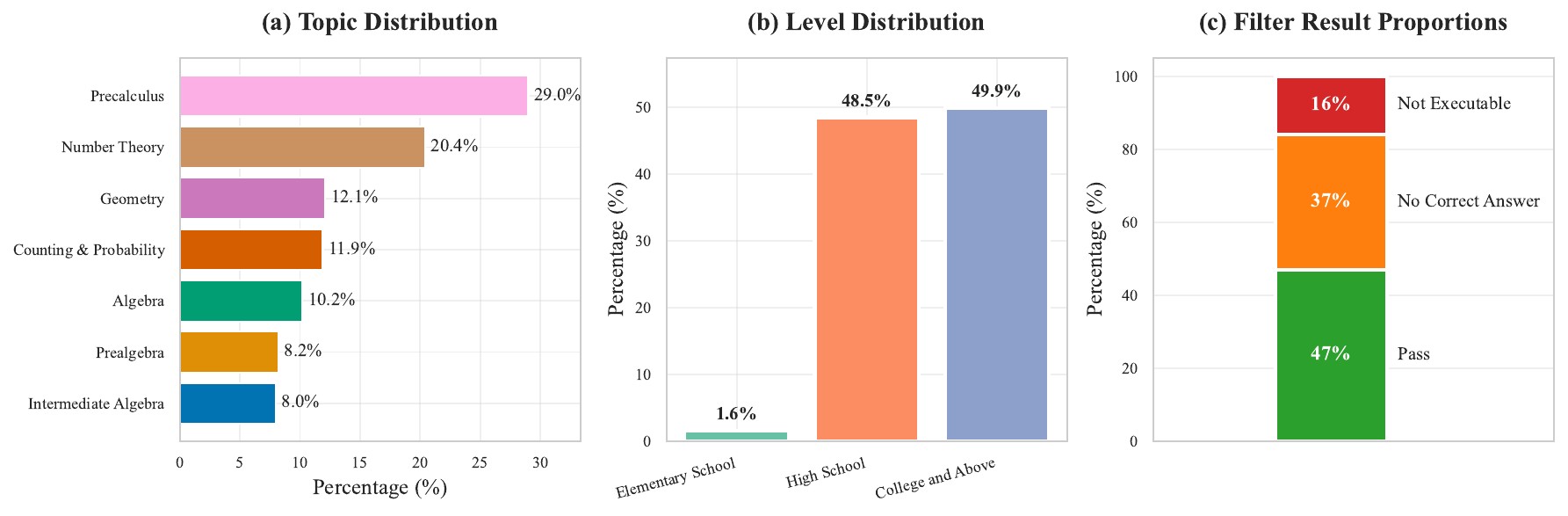}
    \caption{Statistical analysis of the RV-Syn pipeline. (a) Distribution of mathematical topics within the function library. (b) Difficulty distribution of the final synthesized problems. (c) Proportions of data filtered out at different stages of the synthesis process.}
    \label{fig:dataset}
\end{figure*}
Here, we present statistics for the function library and the final dataset.
Figure \ref{fig:dataset} (a) shows the topic distribution in our function library, where each function is labeled according to the categories defined in the MATH dataset.  
Figure \ref{fig:dataset} (b) displays the difficulty distribution of the synthesized dataset, with the majority of samples concentrated in the higher-difficulty range. This distribution helps explain why RV-Syn achieves more substantial performance gains on more challenging benchmarks.  
Figure \ref{fig:dataset} (c) details the filtering proportions during the synthesis pipeline. We observed that 16\% of the generated computational graphs failed to execute and were discarded. Among the remaining valid graphs, an additional 37\% of the generated problem–solution pairs were filtered out because no Chain-of-Thought solution matched the ground-truth answer produced by graph execution. This automated, graph-based quality control mechanism is crucial for ensuring the high correctness of our final dataset.
Table \ref{tab:network_stats} provides additional statistics for the function library.
\begin{table}[htbp]
  \centering
  \small
  \begin{tabular}{lc}
    \toprule
    \small
    \textbf{Metric} & \textbf{Value} \\
    \midrule
    CB Connected Components Number & 23419 \\
    CB Largest Connected Component Size & 10847 \\
    CB Average Degree per Node & 1.225 \\
    TB Connected Components Number & 8 \\
    TB Largest Connected Component Size & 13638 \\
    TB Average Degree per Node & 8607 \\
    \bottomrule
  \end{tabular}
  \caption{Graph-based function library statistics. CB denotes Co-occurrence Based, TB denotes Topic Based.}
  \label{tab:network_stats}
\end{table}

\section{Impact of the Answer Annotation Model }
\label{Annotation}
Previous works often directly fine-tune models on publicly available datasets released by baseline methods and compare their performance. However, we find that the choice of the answer annotation model is one of the most critical factors influencing downstream performance, which introduce substantial unfairness into such comparisons.
To illustrate this, we re-annotate existing datasets using a unified model, Qwen2.5-Math-7B-Instruct, and compare the performance of LLaMA-3-8B-Instruct fine-tuned on 50k these newly annotated datasets versus the original versions. The results in Table \ref{tab:reannotation_results} show that many methods achieve significant performance improvements, and the performance gap between them narrows.
In all our experiments, we use the same annotation model to ensure a fair comparison and to isolate the impact of problem quality. We observe consistent performance improvements from our method. 
Although some of the improvements may appear modest under the uniform annotation setup, they in fact represent substantial enhancements in data quality, which are partially obscured by the controlled evaluation conditions.

\begin{table}[htbp]
  \centering

  \resizebox{0.5\textwidth}{!}{
  \begin{tabular}{lccc}
    \toprule
    \textbf{Dataset} & \textbf{Original} & \textbf{Re-annotation} & \textbf{Performance Change} \\
    \midrule
    Jiuzhang3.0 & 30.5 & 42.1 & +38.0\% \\
    MathScale & 26.5 & 41.9 & +58.1\% \\
    MMIQC & 34.9 & 43.7 & +25.2\% \\
    Numina-Math & 38.7 & 43.9 & +13.4\% \\
    \bottomrule
  \end{tabular}
  }
  \caption{Performance of LLaMA-3-8B-Instruct Fine-tuned on Re-annotated Datasets using Qwen2.5-Math-7B-Instruct.}
  \label{tab:reannotation_results}
\end{table}

\section{Influence of Seed Dataset}
Since RV-Syn utilizes the high-quality Numina-Math dataset as seed data, we determine whether the performance gains stem from our synthesis framework or merely the superior seed data. We conduct a controlled experiment using the Jiuzhang3.0 baseline. We replace its original web-corpus seed data with Numina-Math and train it on 50k synthesized samples.
The results in Table \ref{tab:ablation_seed} show that simply using better seed data with a direct synthesis method (Jiuzhang3.0) does not improve performance; in fact, the score slightly drops compared to its original setting (39.5 vs. 42.1). We attribute this to the fact that direct synthesis methods rely heavily on the diversity of broad pre-training corpora, and constraining them to a structured QA dataset may limit their generative potential. In contrast, RV-Syn achieves a significantly higher average score (44.92), confirming that our improvements are driven by the rational synthesis framework rather than the seed data alone.
\begin{table}[t]
\centering
\small
\begin{tabular}{lcc}
\toprule
\textbf{Method} & \textbf{Seed Data} & \textbf{Avg. Score} \\
\midrule
Jiuzhang3.0 & Original (Web) & 42.1 \\
Jiuzhang3.0 & Numina-Math & 39.5 \\
\textbf{RV-Syn (Ours)} & Numina-Math & \textbf{44.9} \\
\bottomrule
\end{tabular}
\caption{Impact of seed dataset on performance (50k data, LLaMA-3-8B). Comparing our method against Jiuzhang3.0 seeded with Numina-Math.}
\label{tab:ablation_seed}
\end{table}

\begin{table*}[h]

\begin{tabular}{|c|c|}
\hline
\multicolumn{2}{|c|}{Semantic Similarity} \\ 
\hline
\begin{lstlisting}
def find_gcd(a: int, b: int) -> int:
    """
    Find the greatest common divisor of two numbers.
    Args:
    a (int): The first number.
    b (int): The second number.
    Returns:
    int: The GCD of a and b.
    """
    return math.gcd(a, b)
    
\end{lstlisting}
& 
\begin{lstlisting}
def gcd(a, b):  
    """
    Find the greatest common divisor of two numbers.
    Args:
    a (int): The first number.
    b (int): The second number.
    Returns:
    int: The GCD of a and b.
    """
    while b:
        a, b = b, a % b
    return a
    
\end{lstlisting}
\\
\hline
\multicolumn{2}{|c|}{Structure Similarity} \\ 
\hline
\begin{lstlisting}
def calculate_matrix_product(matrix1, matrix2):
    """
    Calculate the product of two matrices.
    Parameters:
        matrix1 (numpy.array): The first matrix.
        matrix2 (numpy.array): The second matrix.
    Returns:
        numpy.array: The product of matrix1 and matrix2.
    """
    return np.dot(matrix1, matrix2)
    
\end{lstlisting}
& 
\begin{lstlisting}
def matrix_transformation(projection_matrix1, projection_matrix2):
    """
    Calculate transformation matrix by multiplying 
    projection matrices.
    Args:
    - projection_matrix1: The first projection matrix.
    - projection_matrix2: The second projection matrix.
    Returns:
        numpy array: The transformation matrix.
    """
    return np.dot(projection_matrix2, projection_matrix1)
    
\end{lstlisting}
\\
\hline
\end{tabular}
\caption{Examples of Semantic and Structure Similarity.}
\label{tab:similar}
\end{table*}

\section{Examples of Function Similarity}\label{sec:simi}
We provide examples of semantic and structural similarity in Table \ref{tab:similar}. Specifically, functions that share these similarities or have the same docstring are grouped into the same function set.


\end{document}